\documentclass[letterpaper, 10 pt, conference]{ieeeconf}

\IEEEoverridecommandlockouts
\usepackage{amsmath}
\usepackage{amssymb}
\usepackage[protrusion=false]{microtype}
\usepackage{tensor}
\usepackage{mathtools}
\usepackage{makecell}
\usepackage{siunitx}
\usepackage{booktabs}

\title{\LARGE \bf
Preintegrated Velocity Bias Estimation to Overcome\\ %
 Contact Nonlinearities in Legged Robot Odometry}

\author{David Wisth, Marco Camurri, Maurice Fallon%
\thanks{The authors are with the Oxford Robotics Institute, Department of
Engineering Science, University
of Oxford, UK. \newline%
\texttt{\{davidw, mcamurri, mfallon\}@robots.ox.ac.uk}}}

\usepackage{amssymb}
\usepackage{todonotes}
\usepackage{algorithm2e}
\usepackage{multirow}

\usepackage{tikz}
\usetikzlibrary{bayesnet}
\usetikzlibrary{arrows,positioning}

\newcommand{\X}{\mathcal{X}}
\newcommand{\Z}{\mathcal{Z}}
\newcommand{\etalcite}[2]{#1~et~al.~\cite{#2}}

\DeclareMathOperator*{\argmax}{arg\,max}
\DeclareMathOperator*{\argmin}{arg\,min}

\newcommand{\hide}[1]{}
\newcommand{\Figure}{Fig.~}
\newcommand{\Equation}{Eq.~}

\newcommand{\eg}{{e.g.,~}}

\newcommand{\bdmath}{\begin{dmath}}
\newcommand{\edmath}{\end{dmath}}
\newcommand{\beq}{\begin{equation}}
\newcommand{\eeq}{\end{equation}}
\newcommand{\bdm}{\begin{displaymath}}
\newcommand{\edm}{\end{displaymath}}
\newcommand{\bea}{\begin{eqnarray}}
\newcommand{\eea}{\end{eqnarray}}
\newcommand{\beal}{\beq \begin{array}{ll}}
\newcommand{\eeal}{\end{array} \eeq}
\newcommand{\beas}{\begin{eqnarray*}}
\newcommand{\eeas}{\end{eqnarray*}}
\newcommand{\ba}{\begin{array}}
\newcommand{\ea}{\end{array}}
\newcommand{\bit}{\begin{itemize}}
\newcommand{\eit}{\end{itemize}}
\newcommand{\ben}{\begin{enumerate}}
\newcommand{\een}{\end{enumerate}}

\newcommand{\SO}{\mathrm{SO}}

\newcommand{\Real}{\mathbb{R}}

\newcommand{\SOthree}{\ensuremath{\SO(3)}\xspace}

\newcommand{\calC}{{\cal C}}

\newcommand{\calI}{{\cal I}}

\newcommand{\calV}{{\cal V}}

\newcommand{\calX}{{\cal X}}
\newcommand{\calZ}{{\cal Z}}

\newcommand{\R}{\mathbf{R}}

\newcommand{\Identity}{\mathbf{I}}

\newcommand{\residual}{\mathbf{r}}
\newcommand{\transpose}{\mathsf{T}}

\newcommand{\rotvel}{\boldsymbol\omega}

\newcommand{\rotvec}{\boldsymbol\phi}

\newcommand{\tran}{\mathbf{p}}

\newcommand{\vel}{\mathbf{v}}

\newcommand{\bias}{\mathbf{b}}

\newcommand{\gravity}{\mathbf{g}}

\newcommand{\expmap}{\mathrm{Exp}}
\newcommand{\logmap}{\mathrm{Log}}

\newcommand{\State}{\boldsymbol{x}}

\newcommand{\World}{\mathtt{W}}
\newcommand{\Imu}{\mathtt{I}}
\newcommand{\Camera}{\mathtt{C}}
\newcommand{\world}{\mathtt{{W}}}

\newcommand{\imu}{\mathtt{{I}}}
\newcommand{\base}{\mathtt{{B}}}
\newcommand{\contact}{\mathtt{{K}}}
\newcommand{\Base}{\mathtt{{B}}}

\newcommand{\foot}{\mathtt{K}}

\newcommand{\meters}{\rm{m}}

\newcommand{\landmark}{\mathbf{m}}

\newcommand{\bv} {\bias^v}

\newcommand{\bvi}{\bias^v_i}
\newcommand{\bwi}{\bias^\omega_i}
\newcommand{\debias}{\delta \bias}
\newcommand{\debiasv}{\delta \bv}
\newcommand{\pbias}{\partial \bias}

\newcommand{\etab}{\boldsymbol{\eta}}

\newcommand{\etav}{\etab^v}
\newcommand{\etavk}{\etab^v_k}

\newcommand{\tv}{\tilde{\vel}}

\newcommand{\tvk}{\tilde{\vel}_k}

\newcommand{\dep}{\delta \tran}
\newcommand{\depij}{\dep_{ij}}

\newcommand{\dpij}{\Delta \tran_{ij}}

\newcommand{\dRik}{\Delta \R_{ik}}
\newcommand{\dPhi}{\delta \rotvec}
\newcommand{\dt}{\Delta t}
\newcommand{\dR}{\Delta \R}
\newcommand{\dtp}{\Delta\tilde{\tran}}

\newcommand{\dtRik}{\Delta\tilde{\R}_{ik}}
\newcommand{\dtpij}{\Delta\tilde{\tran}_{ij}}

\newcommand{\dPhiij}{\delta\rotvec_{ij}}
\newcommand{\dPhiik}{\delta\rotvec_{ik}}

\newcommand{\sumjmo} [1]{\sum_{k=i}^{j-1}\left[ #1 \right]}

\newcommand{\Exp}[1]{\expmap \left( #1 \right)}
\newcommand{\Log}[1]{\logmap \left( #1 \right)}

\newcommand{\defeq}{\triangleq}

\makeatletter
\let\NAT@parse\undefined
\makeatother
\usepackage{hyperref}
\begin{document}

\maketitle
\thispagestyle{empty}
\pagestyle{empty}

\begin{abstract}
In this paper, we present a novel factor graph formulation to estimate the pose
and velocity of a quadruped robot on slippery and deformable terrain. The factor
graph introduces a preintegrated velocity factor that incorporates velocity
inputs from leg odometry and also estimates related biases. From our
experimentation we have seen that it is difficult to model uncertainties at the
contact point such as slip or deforming terrain, as well as leg flexibility. To
accommodate for these effects and to minimize leg odometry drift, we extend the
robot's state vector with a bias term for this preintegrated velocity factor.
The bias term can be accurately estimated thanks to the tight fusion of the
preintegrated velocity factor with stereo vision and IMU factors, without which
it would be unobservable. The system has been validated on several scenarios
that involve dynamic motions of the ANYmal robot on loose rocks, slopes and
muddy ground. We demonstrate a 26\% improvement of relative pose error compared
to our previous work and 52\% compared to a state-of-the-art proprioceptive
state estimator.
\end{abstract}

\section{Introduction}
\label{sec:introduction}

The increased maturity of legged robotics has been demonstrated by the initial
industrial deployments of quadruped robots, as well as impressive results
achieved by academic research. State estimation plays a key role in field
deployment of legged machines: without an accurate estimate of its location and
velocity, the robot cannot build a useful representation of its environment or
plan/execute trajectories to reach goal positions.

Most legged robots are equipped with a  high frequency (\SI{>250}{\hertz})
proprioceptive state estimator for control and local mapping purposes. These are
typically implemented as nonlinear filters fusing high frequency signals such as
kinematics and IMU \cite{Bloesch2013}. In ideal conditions (i.e. high friction,
rigid terrain), these estimators have a limited (yet unavoidable) drift that is
acceptable for local mapping and control.

However, deformable terrains, leg flexibility and slippage can degrade the
estimation performance up to a point where local terrain reconstruction is
unusable and multi-step trajectories cannot be executed, even over short ranges.
This problem is more evident when a robot is moving dynamically.

Recent works have attempted to improve kinematic-inertial estimation accuracy by
detecting unstable contacts and reducing their influence on the overall
estimation \cite{Camurri2017,Jenelten2019}. Alternatively, some works have
focused on incorporating additional exteroceptive sensing modalities into the
estimator to help reduce the pose error \cite{Hartley2018a}.

\begin{figure}
 \centering
 \includegraphics[width=0.8\columnwidth]{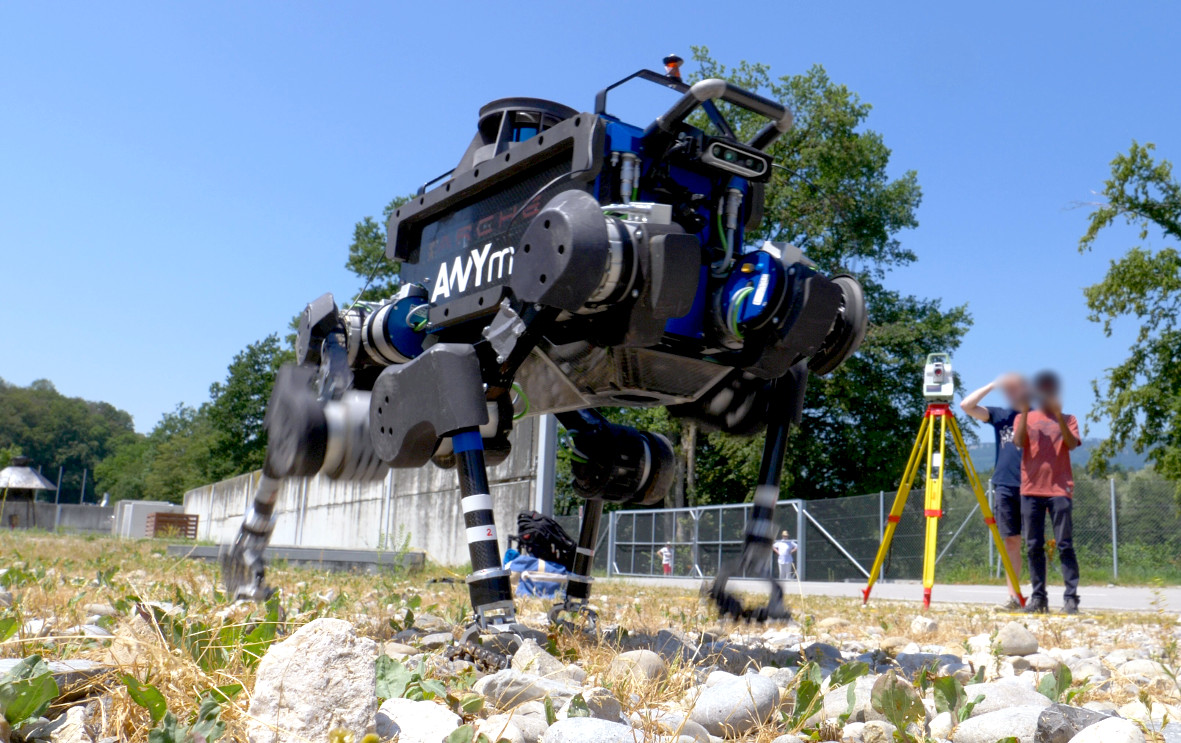}
 \caption{ANYmal trotting over a field of small rocks at the Swiss Military
Rescue Centre in Wangen an der Aare (Switzerland). The ground truth was
collected using a Leica TS16 laser tracker (visible in the background).\newline
Video: \url{http://youtu.be/w1Sx6dIqgQo}}
\label{fig:anymal-leica}
\vspace{-3mm}
\end{figure}

These approaches model the contact locations as being fixed and affected only by
Gaussian noise. Both assumptions fail in conditions such as non-rigid terrain,
kinematic chain flexibility, and foot slippage.

\subsection{Motivation}
\label{subsection:Motivation}
\begin{figure}
\vspace{3mm}
 \centering
 \includegraphics[width=0.8\columnwidth]{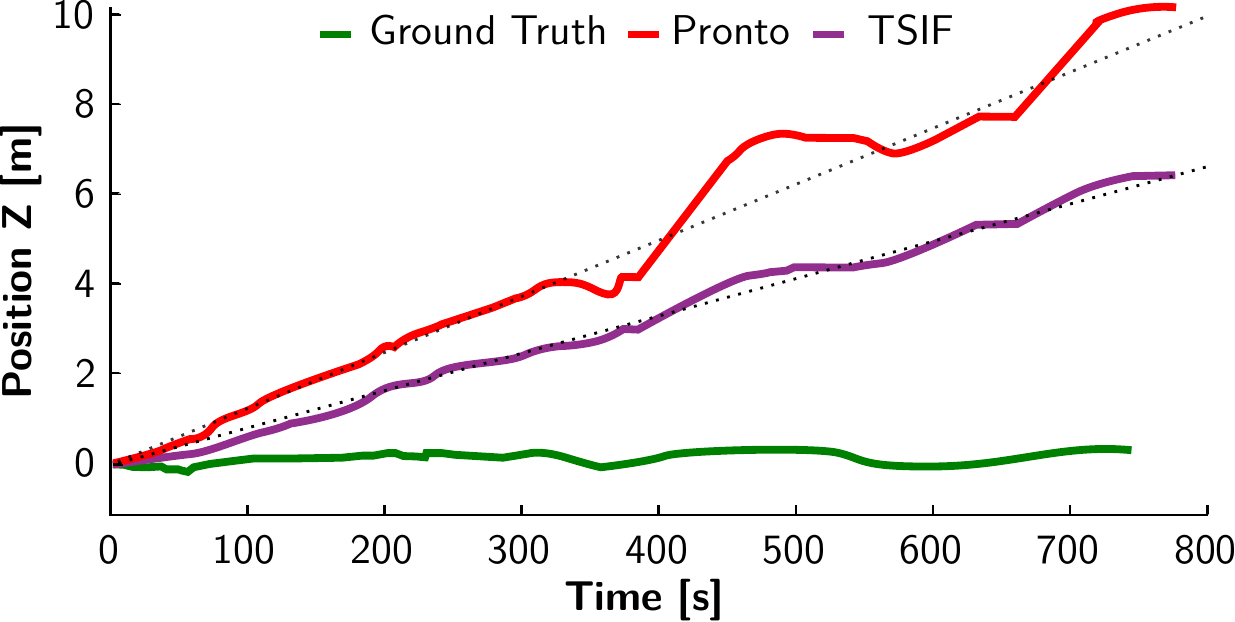}
 \caption{Comparison between estimated robot altitude by Pronto
\cite{Camurri2017} (blue) and TSIF \cite{Bloesch2017} (magenta)
kinematic-inertial state estimators, against ground truth (green) on the SMR1
dataset. Despite local fluctuations, the drift has a characteristic linear
growth.}
 \label{fig:position-drift-motivation}
 \vspace{-6mm}
\end{figure}

Our work is motivated by the observation that there is an
approximately constant velocity bias from the kinematic-inertial state estimator
on the ANYmal robot during dynamic locomotion. An example is shown in Fig.
\ref{fig:position-drift-motivation}, where the robot's estimated altitude
grows linearly as the robot moves.
We attribute this behavior to the
compression of legs and the ground during the contact events.

One approach would be to further model the dynamic properties of the robot
\cite{Koolen2016} or the terrain directly within the estimator. However, this is
likely to be robot specific and terrain dependent: improving performance in one
situation but degrading it elsewhere. Instead, we propose to extend the state of
the estimator with a velocity bias term which is estimated using vision and then
to reject all such effects.

Inspired by the IMU bias estimation and preintegration methods from
\cite{Forster2017}, we propose a novel leg odometry factor that performs online
velocity preintegration and bias estimation to compensate for characteristic
drift in leg odometry.

This factor was implemented as a concurrent thread within our VILENS framework
\cite{Wisth2019}, a visual-inertial-legged estimator which uses GTSAM for
optimization \cite{Dellaert2012}. Thanks to forward propagation, the thread can
output the best pose and velocity estimates at \SI{400}{\hertz} directly, or
just update the bias terms of the estimator running inside the robot's control
loop. The optimized estimate is available from the optimizer thread at at
\SI{30}{\hertz} for effective local mapping.

\subsection{Contribution}
This paper builds upon state-of-the-art methods for online IMU bias estimation
\cite{Forster2017} and the authors' previous work \cite{Wisth2019} to improve
state estimation performance of legged robots in a variety of difficult
scenarios where kinematic-inertial estimates would drift significantly. Compared
to previous research, we present the following contributions:
\begin{itemize}
  \item We present a novel factor graph approach that tightly fuses leg 
        odometry as velocity constraints (as opposed to position 
        constraints), with stereo vision and IMU measurements;
  \item We present the first visual-inertial-legged odometry solution that
        explicitly accounts for error in leg odometry (that can be caused by 
        terrain/leg deformation and slippage) by extending the state with 
        a velocity bias term;
  \item We show that estimating leg odometry error can reduce RPE by 26\% in
        extensive outdoor experiments on muddy ground, slopes, and 
        rockbeds with the ANYmal robot (\Figure \ref{fig:anymal-leica}).
\end{itemize}

The remainder of this article is presented as follows: in Section 
\ref{sec:related-work} we review the literature on mobile state estimation with 
a focus on challenging, outdoor conditions; Section \ref{sec:problem-statement} 
formally defines the problem addressed by the paper and provides the required 
mathematical background; Section \ref{sec:twist-factors} describes the factors
used in our proposed formulation; Section \ref{sec:implementation} presents the 
implementation details of our physical system; Section \ref{sec:results}
presents the experimental results and their interpretation; Section 
\ref{sec:conclusions} concludes with final remarks.

\section{Related Work}
\label{sec:related-work}
In legged robotics, slippage and/or deformation have been typically addressed by
assuming the contact location of a stance foot is always static throughout the
stance period (yet affected by Gaussian noise). Thus, the main focus has been on
detecting and ignoring the feet that are not in fixed contact with the ground.
These methods would typically perform filtering using only proprioceptive
sensing, with a few exceptions.

\etalcite{Bloesch}{Bloesch2013} proposed an Unscented Kalman Filter design that
fuses IMU and differential kinematics. The approach used a threshold on the
Mahalanobis distance of the filter innovation to infer outliers which were then
ignored.

\etalcite{Ma}{Ma2016} proposed an Extended Kalman Filter (EKF) design that was 
Visual Odometry (VO) driven. They incorporated kinematics only when VO failed
or a simple heuristic criteria was met (e.g., when roll or pitch are greater
than \SI{45}{\degree} slippage was assumed and leg odometry ignored). Using 
high-grade sensors, they were able to achieve 1\% error over several kilometers 
of experiments.

\etalcite{Camurri}{Camurri2017} proposed an EKF fusing IMU and differential
kinematics similar to \cite{Bloesch2013}. Instead of the Mahalanobis distance on
the filter innovation, they developed probabilistic contact and impact
detectors. The contact detector learns the optimal force threshold to detect a
foot in contact for a specific gait, while the impact detector adapts the
measurement covariance to reject unreliable measurements. We used this approach
to fuse each leg's kinematic measurements into a single velocity measurement for
our proposed factor graph method.

Recently, \etalcite{Jenelten}{Jenelten2019} presented a probabilistic contact
and slip detector which used a Hidden Markov Model for the ANYmal quadruped
robot. Using differential kinematics, the authors were able to successfully
detect slippage events and robustify locomotion on slippery surfaces. However,
they did not address pose estimate drift.

\section{Problem Statement}
\label{sec:problem-statement}
Our quadruped robot has 12 active Degrees-of-Freedom (DoF) and is equipped with
a stereo camera, an IMU, joint encoders and torque sensors (see Table
\ref{tab:specs} for the specifications). We aim to estimate the history of the
robot's base link pose and its velocity (linear and angular) over time. In
contrast to previous works, we propose to estimate velocity biases (in addition
to IMU biases) to compensate for leg odometry drift, as detailed in the
following section.

\begin{table}
\vspace{3mm}
\centering
\resizebox{\columnwidth}{!}{%
\begin{tabular}{lcrl}
\toprule
\textbf{Sensor} & \textbf{Model} & \textbf{\si{\hertz}} & \textbf{Specs} \\
\midrule
\multirow{2}{*}{IMU} & \multirow{2}{*}{Xsens MTi-100} & \multirow{2}{*}{400} &
\textit{Init Bias:} %
\SI{0.2}{\degree\per\second} $\vert$  \SI{5}{\milli\gram}  \\
 &  &  & \textit{Bias Stab:} %
\SI{10}{\degree\per\hour} $\vert$  \SI{15}{\milli\gram}  \\
\midrule
\multirow{3}{*}{\makecell{Stereo \\ Camera}} &
\multirow{3}{*}{RealSense D435i} & \multirow{3}{*}{30}
 & \textit{Resolution}: $848 \times 480$ px \\
& & &  \textit{FoV:} \SI{91.2 x 65.5}{\degree} \\
& & & \textit{Imager:} IR global shutter \\
\midrule
Encoder & ANYdrive & 400 & \textit{Resolution:} \SI{<0.025}{\degree} \\
\midrule
Torque & ANYdrive & 400 & \textit{Resolution:} \SI{<0.1}{\newton\meter} \\
\bottomrule
\end{tabular}}
\caption{}
\label{tab:specs}
\vspace{-9mm}
\end{table}

\begin{figure}
\centering
\vspace{2mm}
\includegraphics[height=4.5cm]{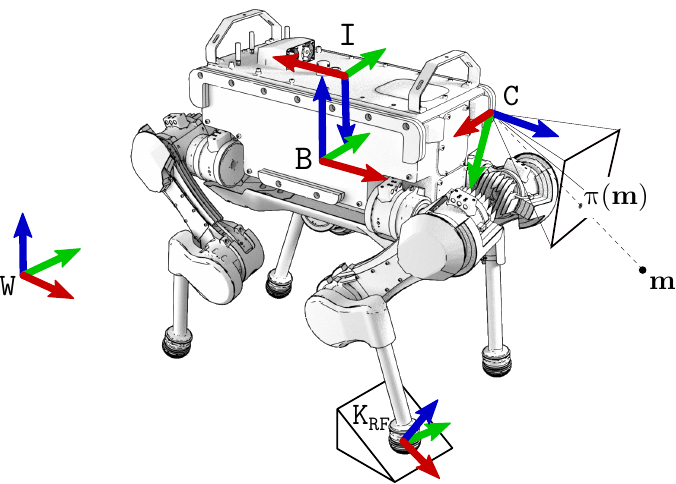}
\caption{Reference frames conventions. The world frame $\World$ is fixed to
earth, while the base frame $\Base$, the camera's optical frame $\Camera$, and
the IMU frame, $\Imu$ are rigidly attached to the robot's chassis. When a foot
touches the ground (\eg the Right Front, RF), a contact frame $\foot$
(perpendicular to the ground and parallel to $\World$'s $y$-axis) is defined.
The projection of a landmark point $\mathbf{m}$ onto the image plane is
$\pi(\mathbf{m})$.}
\label{fig:coordinate-frames}
\vspace{-3mm}
\end{figure}

The relevant reference frames are specified in Fig. \ref{fig:coordinate-frames}
and include: the left camera frame $\Camera$, the IMU frame $\Imu$, the
fixed-world frame $\World$, and the base frame $\Base$. When a foot is in
contact with the ground, a contact frame $\contact$ is also defined.

Unless otherwise specified, position $\tensor[_\world]{\tran}{_{\world\base}}$
and orientation $\R_{\world\base}$ of the base are expressed in world
coordinates, velocities of the base $\tensor[_\base]{\vel}{_{\world\base}},
\tensor[_\base]{\rotvel}{_{\world\base}}$ are in base coordinates (see
\cite{furgale2014notation}), IMU biases
$\tensor[_\imu]{\bias}{^{g}},\;\tensor[_\imu]{\bias}{^{a}}$ are expressed in the
IMU frame, and the velocity biases $\tensor[_\base]{\bias}{^{\omega}},
\tensor[_\base]{\bias}{^{v}}$ are expressed in the base frame.

\subsection{State Definition}
\noindent The robot state at time $t_i$ is defined as follows:
\beq
\State_i \triangleq \left[\R_i,\tran_i,\vel_i,\bias_i \right]
\eeq
where: $\R_i \in \SOthree$ is the orientation,  $\tran_i \in \Real^3$ is
the position, $\vel_i \in \Real^3$ is the linear velocity. The bias vector 
$\bias_i$ is composed as follows:
\beq
\bias_i = [\bias^{g}_i \,\; \bias^{a}_i,\; \bias_i^{\omega},\;
\bias_i^{v}]
\in \Real^{12}
\label{eq:biases}
\eeq
where the first two elements are the usual IMU gyro and accelerometer
biases, and the last two $[ \bias_i^{\omega},\;%
\bias_i^{v} ] = \bias^{Q}_i $ are our proposed angular/linear
velocity biases from leg odometry.

This builds upon the formulation from our previous work \cite{Wisth2019}, 
by incorporating a preintegrated velocity factor (as opposed to a
relative pose factor where the integration was operated by an external filter).

In addition to the robot state, we estimate the position of all observed visual
landmarks $\landmark_{\ell}$. The objective of our estimation problem is then
the union of all the robot states and landmarks visible up to the current time
$t_k$:
\beq
\X_k \defeq \bigcup_{\forall i \in \mathsf{K}_k} \left[ \lbrace \State_i
\rbrace,
\bigcup_{\forall \ell \in \mathsf{M}_i} \lbrace \landmark_{\ell} \rbrace
\right]
\eeq
where $\mathsf{K}_k, \mathsf{M}_i$ are the lists of all the keyframe indices up
to time $t_k$ and  all landmark indices visible at time $t_i$, respectively.

\subsection{Measurements Definition}
For each new stereo camera frame $\calC_i$, collected at time $t_i$, we receive
a number of IMU measurements $\calI_{ij}$ collected between $t_i$ and $t_j$. We
also define $\calV_{ij}$ as the angular and linear velocity measurements from an
external source of leg odometry (Pronto or TSIF). This source would account for
the fusion of multiple legs in contact into one velocity measurement per joint
state measurement. The set of all measurements up to time $t_k$ is therefore
defined as:
\beq
\calZ_k \defeq \bigcup_{\forall i \in \mathsf{K}_k} \lbrace \calC_i,
\calI_{ij}, \calV_{ij} \rbrace
\eeq

\subsection{Maximum-a-Posteriori Estimation}
We wish to maximize the likelihood  of the measurements $\calZ_k$ given the
history of states $\calX_k$:
\begin{equation}
 \X^*_k = \argmax_{\X_k} p(\X_k|\Z_k) \propto
p(\X_0)p(\Z_k|\X_k)
\label{eq:posterior}
\end{equation}
As the measurements are formulated as conditionally independent and corrupted by
white Gaussian noise, \Equation (\ref{eq:posterior}) can be formulated as a
least squares minimization problem:
\begin{multline}
\X^{*}_k = \argmin_{\X_k}
\|\mathbf{r}_0\|^2_{\Sigma_0} 
+  \sum_{i \in \mathsf{K}_k} \|\mathbf{r}_{\calI_{ij}}\|^2_{\Sigma_{\calI_{ij}}} 
+
\sum_{i \in \mathsf{K}_k} \| \mathbf{r}_{\calV_{ij}} \|^2_{\Sigma_{\calV_{ij}}}
\\
+ \sum_{i \in \mathsf{K}_k}\|\mathbf{r}_{\bias_{ij}}\|^2_{\Sigma_{\bias}}
+ \sum_{i \in \mathsf{K}_k} \sum_{\ell \in \mathsf{M}_i} 
\|\mathbf{r}_{i,\mathbf{m}_{\ell}} \|^2_{\Sigma_{\mathbf{m}}}
\end{multline}
where each term is the residual associated to a factor type, weighted by its
covariance matrix; specifically the residuals are: state prior, IMU, velocity,
biases and landmarks.

\begin{figure}
 \centering
 \begin{tikzpicture}
\clip(-4.4,-1.5) rectangle (4,2);
%
\definecolor{color1}{RGB}{57,106,177};
\definecolor{color2}{RGB}{218,124,48};
\definecolor{color3}{RGB}{204,37,41};
\definecolor{color4}{RGB}{62,150,81};
\definecolor{color5}{RGB}{107,76,154};
%
\node[latent, xshift=-3cm,yshift=0cm] (X_0) {$\boldsymbol{x}_0$};
\node[latent, xshift=-1cm,yshift=0cm] (X_1) {$\boldsymbol{x}_1$};
\node[latent, xshift= 1cm,yshift=0cm] (X_2) {$\boldsymbol{x}_2$};
\node[latent, xshift= 3cm,yshift=0cm] (X_3) {$\boldsymbol{x}_3$};
\node[latent, xshift= 4.5cm,yshift=0cm, draw=none, fill=none] (X_4) {};
%
\node[obs, xshift=-2cm,yshift=1.5cm]  (L_0) {$\mathbf{m}_0$};
\node[obs, xshift=0cm, yshift=1.5cm]  (L_i) {$\mathbf{m}_\ell$};
\node[obs, xshift=2cm, yshift=1.5cm]  (L_N) {$\mathbf{m}_N$};
\node[obs, xshift=3.5cm, yshift=1.5cm, draw=none, fill=none] (L_Np1) {};
%
\path (L_0) -- node[auto=false]{\ldots} (L_i);
\path (L_i) -- node[auto=false]{\ldots} (L_N);
\path (X_3) -- node[auto=false]{\ldots} (X_4);
\path (L_N) -- node[auto=false]{\ldots} (L_Np1);
%
\node[midway,circle,draw,fill=color5,xshift=-4cm,yshift=0cm,scale=0.5]
(factor_X0) {};
\draw[thick, color5] (X_0) -- (factor_X0);
\node[midway,circle,draw,fill=none,draw=none,xshift=-3.8cm,yshift=-0.5cm,
scale=0.7] (label_imu) {State Prior};
%
\node[midway,circle,draw,fill=color2,draw=color2,xshift=-2cm,yshift=0cm,
scale=0.5] (factor_imu0) {};
\node[midway,circle,draw,fill=color2,draw=color2,xshift=0cm,yshift=0cm,
scale=0.5] (factor_imu1) {};
\node[midway,circle,draw,fill=color2,draw=color2,xshift=2cm,yshift=0.2cm,
scale=0.5] (factor_imu2) {};
\draw[thick, color2] (X_0) -- (factor_imu0);
\draw[thick, color2] (X_1) -- (factor_imu0);
\draw[thick, color2] (X_1) -- (factor_imu1);
\draw[thick, color2] (X_2) -- (factor_imu1);
\draw[thick, color2] (X_2) -- (factor_imu2);
\draw[thick, color2] (X_3) -- (factor_imu2);
\node[midway,circle,draw,fill=none,draw=none,xshift=-2cm,yshift=0.20cm,
scale=0.7]	(label_imu) {IMU};
%
%
\draw [thick, color4] (X_0) -- +(L_0)
node[midway,circle,draw,fill=color4,scale=0.5] {};
\draw [thick, color4] (X_0) -- +(L_i)
node[near end,circle,draw,fill=color4,scale=0.5] {};
\draw [thick, color4] (X_1) -- +(L_0)
node[near start,circle,draw,fill=color4,scale=0.5] {};
\draw [thick, color4] (X_1) -- +(L_i)
node[midway,circle,draw,fill=color4,scale=0.5] {};
\draw [thick, color4] (X_2) -- +(L_i)
node[midway,circle,draw,fill=color4,scale=0.5] {};
\draw [thick, color4] (X_2) -- +(L_N)
node[midway,circle,draw,fill=color4,scale=0.5] {};
\draw [thick, color4] (X_3) -- +(L_N)
node[midway,circle,draw,fill=color4,scale=0.5] {};
\node[midway,circle,draw,fill=none,draw=none,xshift=-3.4cm,yshift=0.75cm,
scale=0.7] (label_visual) {Stereo Visual};
%
\node[latent, xshift=-3cm,yshift=-1cm] (B_0)  {$\boldsymbol{b}_0^Q$};
\node[latent, xshift=-1cm,yshift=-1cm] (B_1)  {$\boldsymbol{b}_1^Q$};
\node[latent, xshift= 1cm,yshift=-1cm] (B_2)  {$\boldsymbol{b}_2^Q$};
\node[latent, xshift= 3cm,yshift=-1cm] (B_3)  {$\boldsymbol{b}_3^Q$};
\node[latent, xshift= 4.5cm,yshift=-1.5cm, draw=none, fill=none] (B_4) {};
%
\node[midway,circle,draw,fill=color3,draw=color3,xshift=-2cm,yshift=-0.5cm,
scale=0.5] (factor_lo0) {};
\node[midway,circle,draw,fill=color3,draw=color3,xshift=0cm,yshift=-0.5cm,
scale=0.5] (factor_lo1) {};
\node[midway,circle,draw,fill=color3,draw=color3,xshift=2cm,yshift=-0.2cm,
scale=0.5] (factor_zvpose) {};
\node[midway,circle,draw,fill=color3,draw=color3,xshift=2cm,yshift=-1cm,
scale=0.5] (factor_zvbias) {};
\draw[thick, color3] (X_0) -- (factor_lo0);
\draw[thick, color3] (B_0) -- (factor_lo0);
\draw[thick, color3] (X_1) -- (factor_lo0);
\draw[thick, color3] (B_1) -- (factor_lo0);
\draw[thick, color3] (X_1) -- (factor_lo1);
\draw[thick, color3] (B_1) -- (factor_lo1);
\draw[thick, color3] (X_2) -- (factor_lo1);
\draw[thick, color3] (B_2) -- (factor_lo1);
\draw[thick, color3] (X_2) -- (factor_zvpose);
\draw[thick, color3] (B_2) -- (factor_zvbias);
\draw[thick, color3] (X_3) -- (factor_zvpose);
\draw[thick, color3] (B_3) -- (factor_zvbias);
\node[midway,circle,draw,fill=none,draw=none,xshift=-1cm,yshift=-0.5cm,
scale=0.7] (label_leg) {Preint. Twist};
\node[midway,circle,draw,fill=none,draw=none,xshift=2cm,yshift=-0.5cm,
scale=0.7] (label_leg) {Zero Velocity Mode};
%
\node[midway,circle,draw,fill=color5,xshift=-4cm,yshift=-1cm,scale=0.5]
(prior_b0) {};
\draw[thick, color5] (B_0) -- (prior_b0);
\end{tikzpicture}
 \caption{VILENS factor graph structure, showing initial prior, visual, IMU, and
preintegrated velocity factors. When a zero velocity state is detected (e.g.
between $x_2$ and $x_3$) then the velocity bias is not used and is kept constant
(we assume the bias is only present when the robot is moving).}
\label{fig:factor-graph}
\vspace{-3mm}
\end{figure}
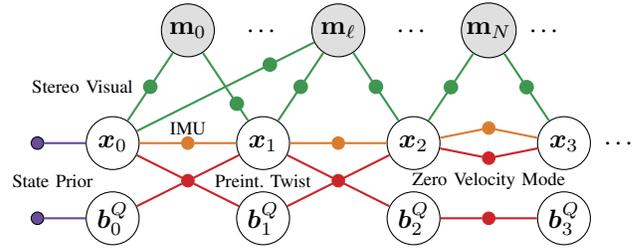

\section{Factor Graph Formulation}
\label{sec:twist-factors}
In the following sections we describe the measurements, residuals and
covariances of the factors which compose the factor graph shown in Fig.
\ref{fig:factor-graph}. For convenience, we summarize the IMU factors from
\cite{Forster2017} in Section \ref{sec:imu-residuals}; our novel velocity factor
is detailed in Section \ref{sec:velocity-factor}; Sections
\ref{sec:bias-residuals} and \ref{sec:stereo-factors} describe the bias and
stereo visual residuals, which are adapted from \cite{Forster2017,Wisth2019} to
include the velocity bias term and stereo cameras, respectively.

\subsection{Preintegrated IMU Factors}
\label{sec:imu-residuals}
In the standard manner, the IMU measurements are preintegrated to constrain the
pose and velocity between two consecutive nodes of the graph, and provide high
frequency state updates between them. This uses a residual of the form:
\begin{equation}
\mathbf{r}_{\calI_{ij}}  = \left[ \mathbf{r}^\transpose_{\Delta
\R_{ij}}, \mathbf{r}^\transpose_{\Delta \vel_{ij}},
\mathbf{r}^\transpose_{\Delta \tran_{ij}} \right]
\end{equation}
where $\calI_{ij}$ are the IMU measurements between $t_{i}$ and $t_j$. The
individual elements of the residual are defined as:
\begin{align}
\mathbf{r}_{\Delta \R_{ij}}   &= \Log{ \Delta\tilde{\R}_{ij}(\bias^g_{i}) } 
\R^\transpose_{i} \R_j \label{eq:rot-residual}\\
\mathbf{r}_{\Delta \vel_{ij}}  &=   \R^\transpose_{i} \left( \vel_j -
\vel_{i} - \gravity \Delta t_{ij} \right) - \Delta 
\tilde{\vel}_{ij}(\bias^g_{i}, \bias^a_{i})\\
\mathbf{r}_{\Delta \tran_{ij}} &=  \R^\transpose_{i} \left( \tran_j -
\tran_{i} - \vel_{i}\Delta t_{ij} - \frac{1}{2}\gravity \Delta
t_{ij}^2 \right) \nonumber\\
&\hspace{3cm}- \Delta \tilde{\tran}_{ij}(\bias^g_{i}, \bias^a_{i})
\end{align}
for the definition of the preintegrated IMU measurements $\Delta\tilde{\R}_{ij},
\Delta \tilde{\tran}_{ij}, \Delta \tilde{\vel}_{ij} $, the noise terms
$\dPhi,\delta\vel,\delta\tran$, and the covariance matrix $\Sigma_{\calI_{ij}}$,
the reader is invited to consult \cite{Forster2017}.

\subsection{Preintegrated Velocity Factors}
\label{sec:velocity-factor}
\subsubsection{Leg Odometry}
When a leg is in rigid, non-slipping contact with the ground, the robot's linear
velocity can be computed from the foot velocity and position in base frame:
\beq
\tensor[_{\base}]{\vel}{_{\world\base}} =
{-\tensor[_\base]{\vel}{_{\base\contact}}} -
\tensor[_\base]{\boldsymbol{\omega}}{_{\world\base}} \times
{\tensor[_\base]{\tran}{_{\base\contact}}}
\label{eq:no-slip}
\eeq
From the sensed joint positions and velocities
$\tilde{\mathbf{q}},\tilde{\dot{\mathbf{q}}}$ and noise
$\boldsymbol{\eta}^q,\boldsymbol{\eta}^{\dot{q}}$ we can rewrite \Equation
(\ref{eq:no-slip}) as a linear velocity measurement \cite{Bloesch2013}:
\beq
\tv = -
J(\tilde{\mathbf{q}}-\boldsymbol{\eta}^q)\cdot(\tilde{\dot{\mathbf{q}}} -
\boldsymbol{\eta}^{\dot{q}}) - \rotvel \times
\text{f}(\tilde{\mathbf{q}}-\boldsymbol{\eta}^q)
\label{eq:legodo}
\eeq
where $\mathrm{f(\cdot)}$ and $J(\cdot)$ are the forward kinematics function and
its Jacobian, respectively.

Eq. (\ref{eq:legodo}) is valid only when the corresponding leg is in contact
with the ground. However, this happens intermittently while the robot moves.
Since multiple legs can be in contact simultaneously, measurement fusion is
necessary. To do this we take advantage of the contact detection and sensor
fusion features of the EKF filter in \cite{Camurri2017} and use it as an
independent source of unified velocity measurements $\tilde{\vel},
\tilde{\boldsymbol{\omega}}$.

\subsubsection{Velocity Bias}
On slippery/deformable terrains, the constraint from \Equation
(\ref{eq:no-slip}) might not be respected, leading to leg odometry drift and
inconsistency with visual odometry. In our experience this drift is constant and
gait dependent.

For these reasons, we relax \Equation (\ref{eq:no-slip}) by  adding a slowly
varying bias term $\bv$ to \Equation (\ref{eq:legodo}). As in
\cite{Bloesch2013}, we also collect all the effects of encoder noise into a
single term, leading to:
\begin{align}
\tilde{\vel} &= -
J(\tilde{\mathbf{q}})\tilde{\dot{\mathbf{q}}} - \rotvel \times
\text{f}(\tilde{\mathbf{q}}) + \bv + \boldsymbol{\eta}^{v}
\\
\tilde{\rotvel} &= \rotvel + \bias^{\omega} + \boldsymbol{\eta}^{\omega}
\label{eq:omega-measurement}
\end{align}
where the parameters for $\boldsymbol{\eta}^{v}, \boldsymbol{\eta}^{\omega}$ are
provided by the source of velocity measurements.

\subsubsection{Preintegrated Measurements}

In the following, we derive the the preintegrated position and noise only.
For the respective rotational quantities $\Delta\tilde\R$ and $\dPhi$, we refer
to \cite{Forster2017}, as they have the same form as for IMU
measurements.

Assuming constant velocity between $t_i$ and $t_j$, we can iteratively calculate
the position at time $t_j$ as:
\beq
\tran_j  =  \tran_i + \sumjmo{\R_{k} ( \tvk - \bvi - \etav_k) \dt}
\label{eq:iterative-pos}
\eeq
From \Equation (\ref{eq:iterative-pos}) a relative measurement can be obtained:
\begin{equation}
\dpij = \R_i^\transpose(\tran_j-\tran_i) =
\sumjmo{\dR_{ik}(\tvk-\bias^v_k-\etav_k)\dt)}
\label{eq:rel-meas}
\end{equation}

With the substitution $\dRik = \dtRik\Exp{-\dPhiik}$ to include the
preintegrated rotation measurement, and the approximation $\Exp{\rotvec}
\simeq \Identity + \rotvec^{\wedge}$, \Equation
(\ref{eq:rel-meas}) becomes:
\begin{equation}
\dpij \simeq \sumjmo{ \dtRik (I -\dPhi_{ik}^\wedge)(\tvk - \bvi - \etavk)
\dt }
\label{eq:rel-meas-2}
\end{equation}
By separating the measurement and noise components of \Equation
(\ref{eq:rel-meas-2}) and ignoring the higher order terms, we can define the
preintegrated position \textit{measurement} $\dtp$ and \textit{noise} $\dep$ as:
\begin{align}
\dtpij  &\defeq  \sumjmo{ \dtRik (\tvk - \bvi)\dt } \label{eq:rel-meas-3}\\
 \depij  &\defeq  \sumjmo{\dtRik \etavk \dt - \dtRik (\tvk - \bvi)^\wedge
 \dPhiij \dt}
 \label{eq:preint-noise-model}
\end{align}
Note that both quantities still depend on the bias states $\bias^Q =
[\bias^\omega,\bias^v]$; when these change, we would like to avoid the
recomputation of \Equation (\ref{eq:rel-meas-3}).
Given a small change $\delta \bias^Q$ such that $\bias^Q = \bar{\bias}^Q +
\delta \bias^Q$, we approximate the measurement as:
\begin{equation}
\dtpij(\bias^Q)  \simeq  \dtpij(\bar{\bias}^Q) + \frac{\partial
\dpij}{\pbias^\omega}\debias^\omega + \frac{\partial
\dpij}{\pbias^v}\debiasv
\end{equation}

\subsubsection{Residuals}
the velocity residual can be expressed as:
\beq
\residual_{\calV_{ij}} = \left[ \residual_{\dR_{ij}}^\transpose,
\residual_{\Delta
\tran_{ij}}^\transpose \right]
\eeq
where $\residual_{\dR_{ij}}$ has the same form as \Equation 
(\ref{eq:rot-residual}) and
$\residual_{\dpij}$ is:
\beq
\residual_{\dpij} = \R_i^\transpose \left( \tran_j - \tran_i \right) -
\dtpij(\bwi,\bvi)
\eeq

\subsubsection{Covariance}
After simple manipulation of \Equation\ref{eq:preint-noise-model}, the
covariance of the residual $\residual_{\dpij}$ can be expressed as a linear
combination of the preintegrated and current sensor noise:
\beq
\Sigma_{\calV_{ik+1}} = A\; \Sigma_{\calV_{ik}}\; A^\transpose +
B\; \Sigma^\eta_{\calV}\;B^\transpose
\eeq
$\Sigma_{\calV_{ik}}$ evolves over time while $\Sigma^\eta_{\calV}$ is fixed and
depends sensor specifications. The multiplicative terms are:
\begin{equation}
 A = \begin{bmatrix}
      \dtRik^\transpose & 0 \\
      -\dtRik(\tv_k - \bv)^\wedge\dt & \Identity
     \end{bmatrix},
     B = \begin{bmatrix}
     J^k_r\dt & 0 \\
     0 & \dtRik \dt
        \end{bmatrix}
\end{equation}
where $J_r$ is the right Jacobian of $\SOthree$ and the other terms are
manipulations of $\dPhi$ and $\delta\tran$ from \cite{Forster2017} and \Equation
(\ref{eq:preint-noise-model}).

\subsection{Bias Residuals}
\label{sec:bias-residuals}
The biases from \Equation (\ref{eq:biases}) are intended to change slowly and
are therefore modeled as a Gaussian random walk. The residual term for the cost
function is therefore:
\begin{multline}
\|\mathbf{r}_{\bias_{ij}}\|^2_{{\Sigma}_\bias} \defeq
\|{\bias^g_{j}}-{\bias^g_{i}}\|^2_{\Sigma_{\bias^g}}
+\|{\bias^a_{j}}-{\bias^a_{i}}\|^2_{\Sigma_{\bias^a}}+\\
+\|{\bias^\omega_{j}}-{\bias^\omega_{i}} \|^2_{\Sigma_{\bias^\omega}}+\| {
\bias^v_ {j} } -{\bias^v_{i}}\|^2_{\Sigma_{\bias^v}}
\end{multline}
where the covariance matrices are determined by the expected rate of change of
these quantities, depending on the IMU specifications or the drift rate of the
leg odometry.

\subsection{Stereo Visual Factors}
\label{sec:stereo-factors}
Given a stereo pair of rectified images, the stereo visual odometry residual is
the difference between the measured landmark pixel locations $(u^{L}, v),
(u^{R}, v)$, and the re-projection of the estimated landmark location into image
coordinates, $(\pi_u^L, \pi_v),(\pi_u^R, \pi_v)$ using the standard
radial-tangential distortion model. The residual at pose $i$ for landmark
$\landmark_\ell$ is:
\begin{equation}
\mathbf{r}_{i,\mathbf{m}_\ell} =
\left( \begin{array}{c}
\pi_u^L(\R_i,\tran_i, \mathbf{m}_\ell) - u^L_{i,\ell} \\
\pi_u^R(\R_i,\tran_i, \mathbf{m}_\ell) - u^R_{i,\ell} \\
\pi_v(\R_i,\tran_i, \mathbf{m}_\ell) - v_{i,\ell}
\end{array} \right)
\end{equation}
where $\Sigma_\mathbf{m}$ is computed using an uncertainty of \num{0.5} pixels.

\section{Implementation}
\label{sec:implementation}
\begin{figure}
\vspace{3mm}
\centering
\includegraphics[width=0.85\columnwidth]{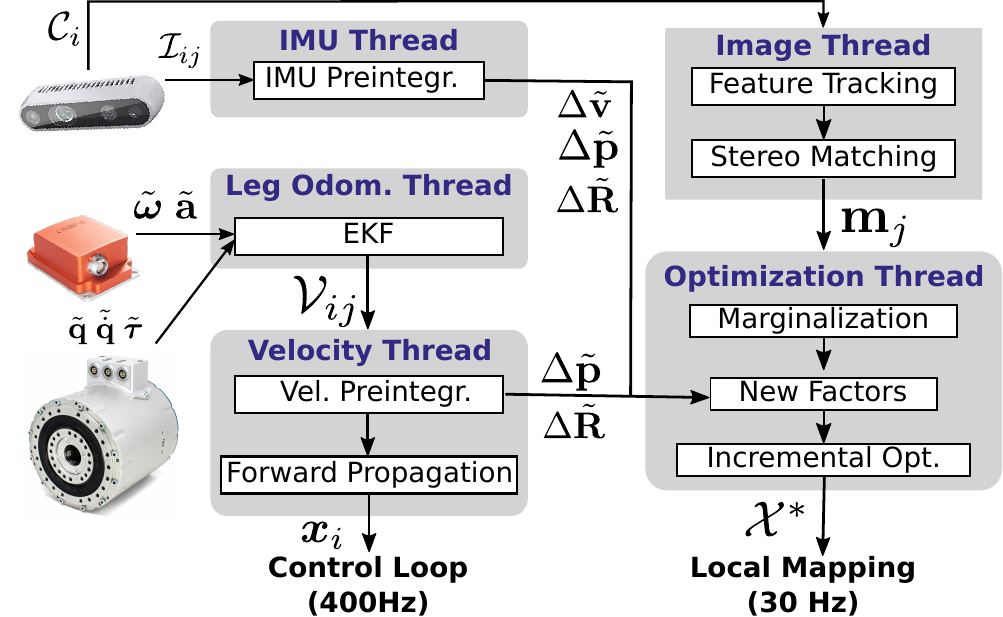}%
\caption{The VILENS architecture with preintegrated velocity bias estimation.}
\label{fig:state-estimation-architecture}
\vspace{-6mm}
\end{figure}

The state estimation architecture is shown in \Figure
\ref{fig:state-estimation-architecture}. Four parallel threads execute the
following operations: preintegration of the IMU factor, preintegration of the
velocity factor, stereo feature tracking, and optimization. This approach
outputs \SI{400}{\hertz} velocity and pose estimates from the preintegration
thread for use by the robot's control system, and a \SI{30}{\Hz} output from the
factor-graph optimization thread for use by local mapping. When a new keyframe
is processed, the preintegrated measurements and tracked landmarks are collected
by the optimization thread, while the other threads process the next set of
measurements.
The factor graph optimization is implemented using the efficient incremental
optimization solver iSAM2 \cite{Kaess2012}, which is part of the GTSAM library
\cite{Dellaert2017}. We limit the number of states in the graph to 500 to
keep the optimization time approximately constant.

\subsection{Visual Feature Tracking}
We detect features using the FAST corner detector, and track them between
successive frames using the KLT feature tracker. Outliers are rejected using a
RANSAC-based rejection method. Thanks to the parallel architecture and
incremental optimization, all frames are used as keyframes, achieving
\SI{30}{\hertz} nominal output. In contrast to \cite{Wisth2019}, we use the
Dynamic Covariance Scaling (DCS) \cite{MacTavish2015} robust cost function to
reduce the effect of landmark correspondence outliers on the optimization.

\subsection{Zero Velocity Update Factors}
To limit drift and factor graph growth when the robot is stationary, we enforce
zero velocity updates on the different sensor modalities (camera, IMU, and leg
odometry). If two out of three modalities report no motion, a zero velocity
constraint factor is added to the graph. The IMU and leg odometry threads report
zero velocity when position (rotation) is less than \SI{0.1}{\milli\meter}
(\SI{0.5}{\degree}) between two keyframes. The image thread reports zero
velocity when less than \SI{0.5}{\text{pixels}} displacement of all the features
is detected over the same period.

\section{Experimental Results}
\label{sec:results}
\begin{figure}
\vspace{3mm}
\centering
\includegraphics[width=0.43\columnwidth]{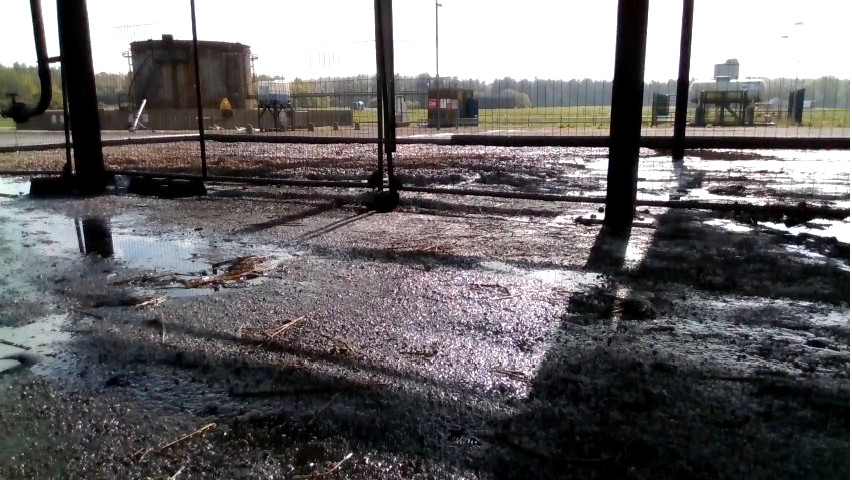}%
\hspace{0.01\columnwidth}%
\includegraphics[width=0.43\columnwidth]{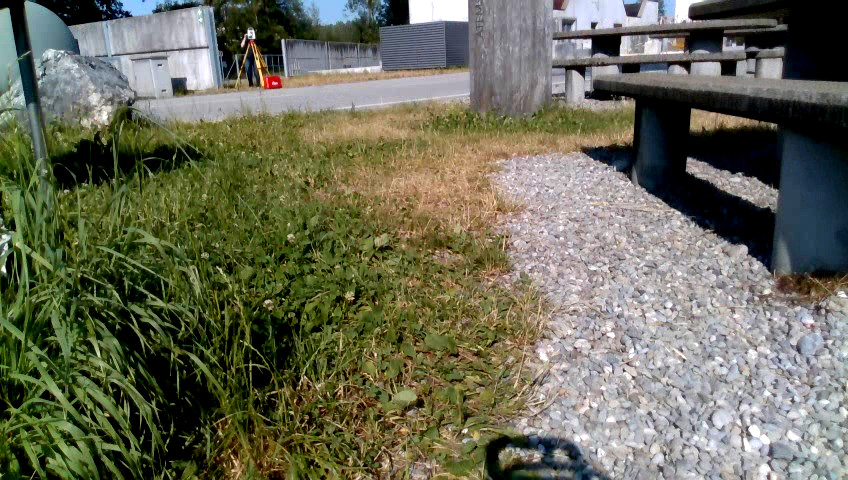}\\
\vspace{0.01\columnwidth}
\includegraphics[width=0.43\columnwidth]{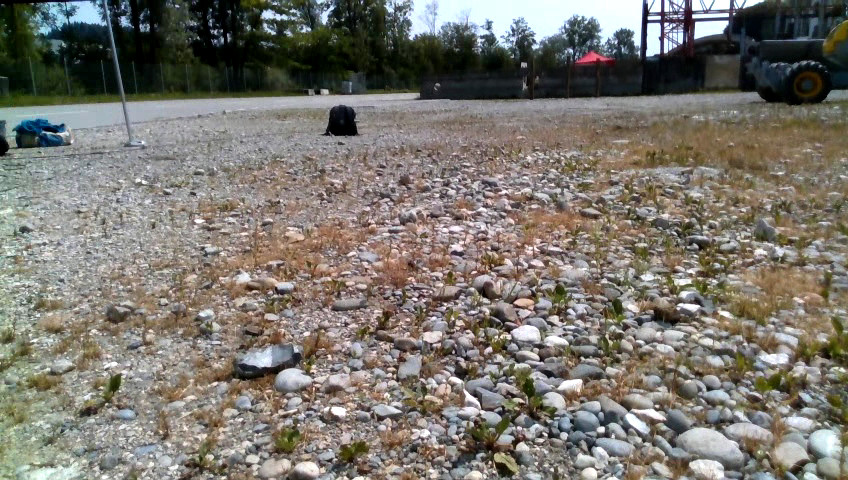}%
\hspace{0.01\columnwidth}%
\includegraphics[width=0.43\columnwidth]{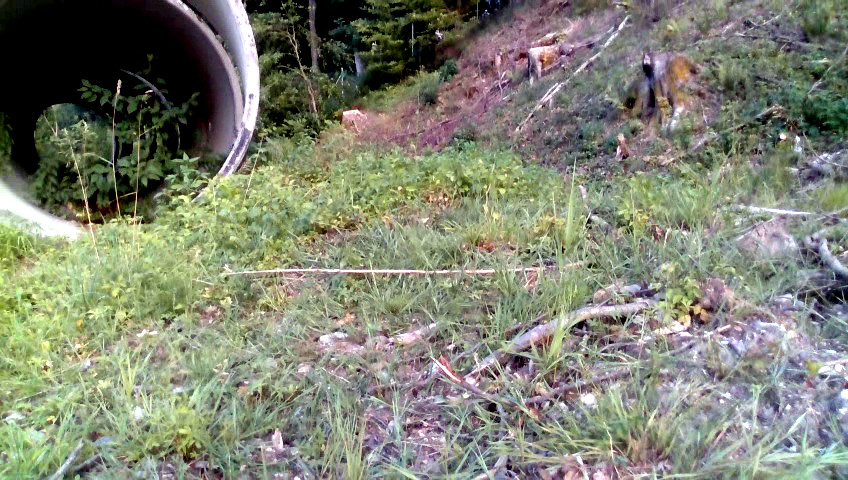}
\caption{Onboard camera feed from the four scenarios evaluated. \textit{Top
left}: wet concrete
(FSC); \textit{Top right:} gravel and grass (SMR1); \textit{Bottom left:}
rockbeds (SMR2); \textit{Bottom right:} muddy grass uphill
(SMR3).}
\label{fig:fsc-scenarios}
\end{figure}

We have tested our proposed algorithm on a variety of terrain types for a total
time of \SI{53}{\minute} and \SI{403}{\meter} traveled distance. The datasets
consist of four scenarios (Fig. \ref{fig:fsc-scenarios}):
\begin{itemize}
\item\textbf{FSC} a \SI{240}{\meter} long trajectory consisting of three loops 
on wet concrete, standing water/oil, gravel and mud;
\item\textbf{SMR1} a \SI{106}{\meter} straight trot over concrete, gravel and
high grass, followed by two loops on short grass alternating between dynamic and
static gaits;
\item\textbf{SMR2} a \SI{22}{\meter} straight trot on rockbeds;
\item\textbf{SMR3} a \SI{35}{\meter} trot in a loop uphill with grass, mud
and external disturbances applied to the robot to cause slippage events.
\end{itemize}
The first dataset was collected at Fire Service College (FSC), Moreton-on-Marsh,
UK; the other three at the Swiss Military Rescue Center (SMR), Wangen an der
Aare, Switzerland. Different copies of the ANYmal robot were used in the
experiments. The attached video gives a sense of the conditions.

To generate ground truth, we tracked the robot using the Leica TS16 laser
tracker (shown in Fig. \ref{fig:anymal-leica}), which provides millimeter
accurate position measurements at \SI{5}{\Hz}. The orientation was reconstructed
with an optimization method similar to that used by the EuRoC
dataset~\cite{Burri2016}.

We have evaluated the Relative Pose Error (RPE) over a distance of
\SI{10}{\meters} for the following algorithms:
\begin{itemize}
 \item \textbf{V-VB:} VILENS with our proposed velocity bias factors;
\item \textbf{V-RP:} VILENS with leg odometry integrated as relative
pose factors, as used in our previous work \cite{Wisth2019};
\item \textbf{V-VI:} a pure visual inertial navigation system (i.e.,
VILENS without leg odometry factors);
\item \textbf{TSIF:} default ANYmal state estimator \cite{Bloesch2017}, our
baseline.
\end{itemize}
Note that the same IMU and camera settings have been used for all configurations
and datasets. Also, in comparison to our previous work \cite{Wisth2019}, the
robustness of visual feature tracking has been improved due to the introduction
of stereo factors, higher framerates and robust cost functions.

\begin{figure}
\vspace{3mm}
 \centering
 \includegraphics[width=0.95\columnwidth]{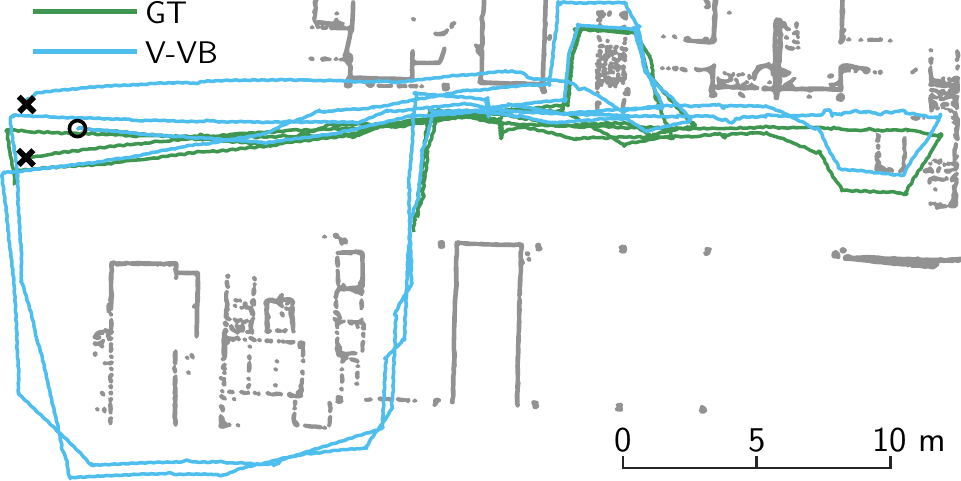}
 \caption{Aerial view of the trajectories for method V-VB and the ground truth
on the FSC dataset (\SI{240}{\meter} traveled). The common start point is
indicated with a circle and the two endpoints with a cross. The lower part of
the trajectory has no ground truth. Note that V-VB has no loop-closure system.}
\label{fig:fsc-trajectory}
\end{figure}

\begin{table}
\centering
\begin{tabular}{l|cccc}
\toprule
 \multicolumn{5}{c}{ \textbf{\SI{10}{\meter} Relative Pose Error (RPE)
$\boldsymbol{\mu
(\sigma)}$
[\si{\metre}]}} \\
\midrule
\textbf{Data} & \textbf{TSIF}\cite{Bloesch2017} &
\textbf{V-VI} & \textbf{V-RP}\cite{Wisth2019} &%
\textbf{V-VB} \\
\midrule
FSC &  0.49 (0.36) & 0.42 (0.32) & 0.47 (0.40) & \textbf{0.36}
(0.30) \\
SMR1 & 0.96 (0.44) & \textbf{0.36} (0.29) & \textbf{0.36} (0.32) & \textbf{0.36}
(0.33) \\
SMR2 & 0.69 (0.23) & \textbf{0.24} (0.16) & 0.45 (0.10) & \textbf{0.24}
(0.16)\\
SMR3 & 0.87 (0.42) & 0.43 (0.48) & 0.53 (0.46) & \textbf{0.39}
(0.48)\\
\bottomrule
\end{tabular}
 \caption{}
\label{tab:rpe}
\vspace{-9mm}
\end{table}
The results are summarized in Table \ref{tab:rpe}. In three of the four datasets
(FSC, SMR2 and SMR3), V-VI is able to outperform V-RP. This is because the
datasets are designed to be particularly challenging for leg odometry.
Therefore, the inclusion of relative pose factors without compensating for leg
odometry drift actually degrades the performance compared to a visual-inertial
only system. With the preintegrated velocity bias estimation, leg odometry
improves the estimate up to \SI{14}{\percent} compared to V-VI and
\SI{26}{\percent} compared to V-RP.

The global performance is shown in \Figure \ref{fig:fsc-trajectory}, which
depicts the estimated and ground truth trajectories on the \SI{240}{\meter} FSC
dataset. By incorporating visual information to reject drift, the final z
position of V-VB is \SI{8.6}{\centi\meter} above ground truth, compared to a
drift of \SI{4.02}{\meter} from the TSIF kinematic-inertial estimator. Note that
since VILENS is an odometry system, no loop closures have been performed.

\subsection{Analysis of Visually Challenging Episodes}
\label{sec:poor-vo}

Most of the datasets presented favorable conditions for VO (well lit
static scenes with texture). However, there were also certain locations where
feature tracking struggled.

\begin{figure}
 \centering
 \includegraphics[height=3cm]{%
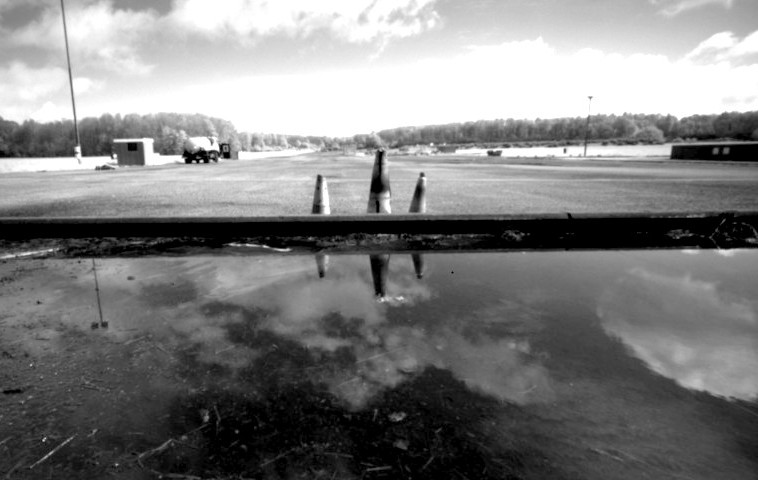}
 \includegraphics[height=3cm]{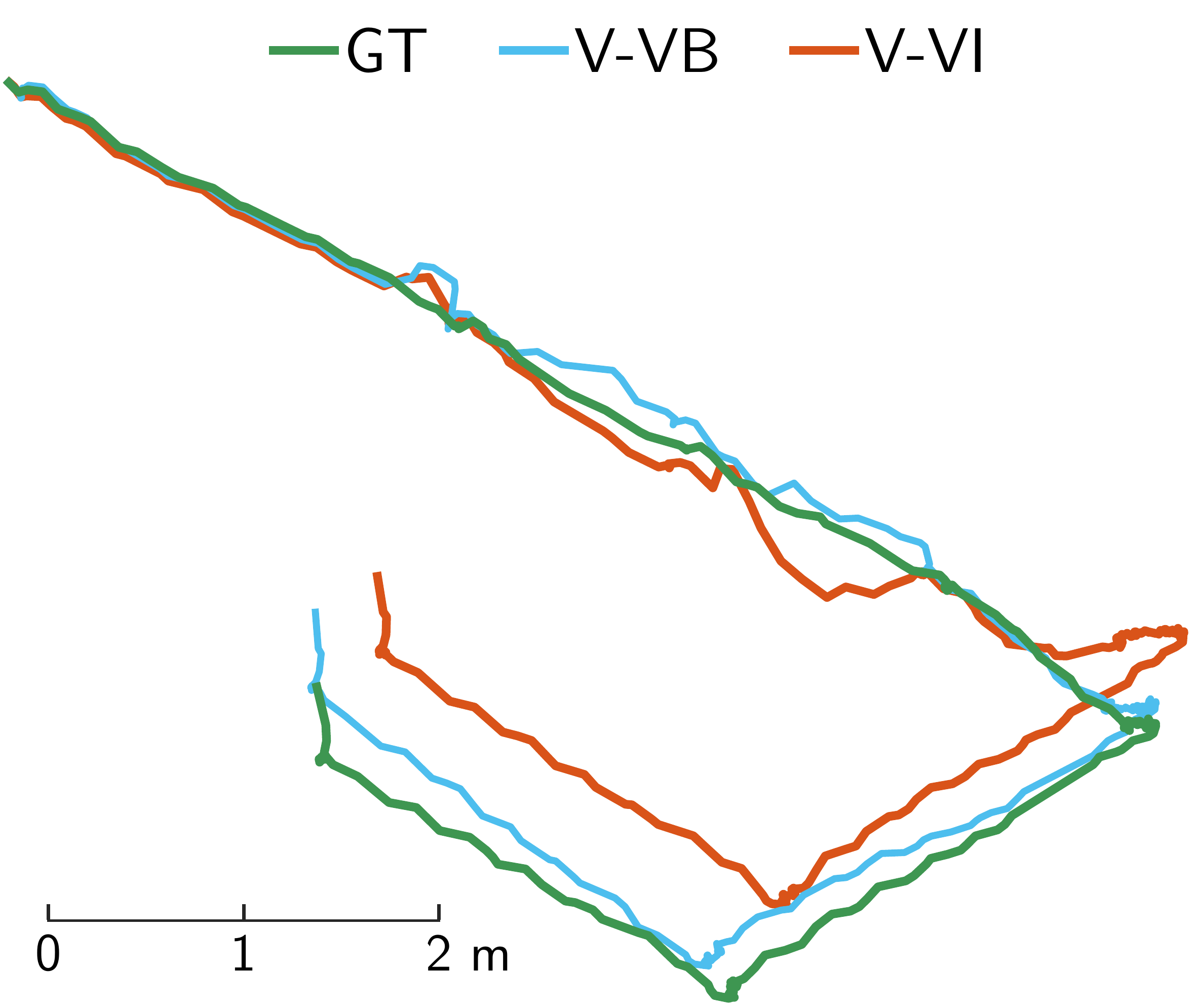}
 \caption{\textit{Left:} Onboard image of a visually challenging scene
containing reflections from a large, oily puddle. \textit{Right:} Top-down
comparison of V-VB and V-VI trajectories aligned with ground truth while
crossing the puddle. }
\label{fig:example_poor_vo}
\vspace{-3mm}
\end{figure}

\Figure \ref{fig:example_poor_vo} shows a situation from FSC dataset where the
robot traverses a large puddle. V-VI  tracks the features on the water, causing
drift in the lateral direction. Instead, V-VB maintains a better pose estimate
by relying on leg odometry, whose drift is suppressed using the estimated
velocity bias.

\subsection{Velocity Bias Evolution}
\label{sec:bias-evo}

\begin{figure}
\vspace{3mm}
 \centering
 \includegraphics[width=0.95\columnwidth]{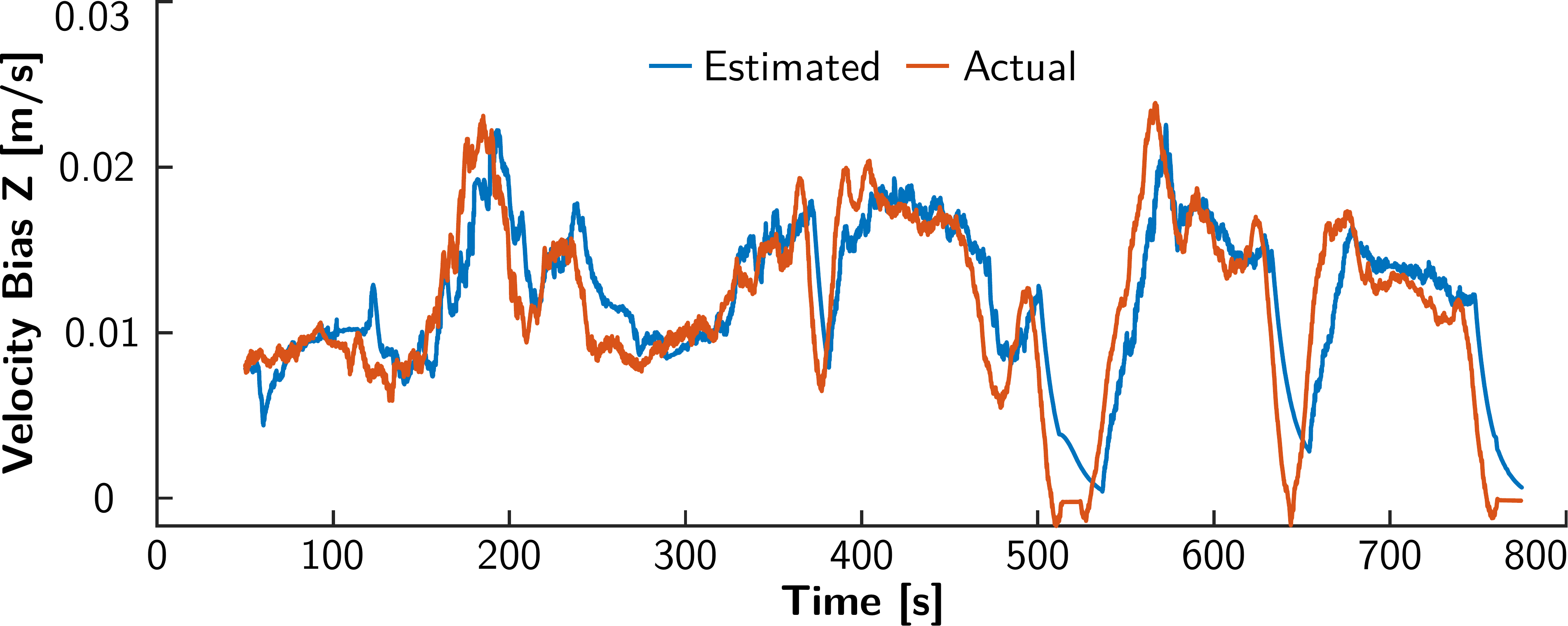}
 \caption{Using visual information, VILENS is able to accurately and stably
estimate the bias of the kinematic-inertial estimate. Experiment: SMR1.}
\vspace{-2mm}
\label{fig:bias-evo}

\end{figure}

We have compared the estimated online bias in the z-axis to a lowpass filtered
version of the same signal from the Pronto EKF \cite{Camurri2017} (\Figure
\ref{fig:bias-evo}). The sequence analyzed is the same as the one shown in
\Figure \ref{fig:position-drift-motivation}. Since the z-axis position and
average velocity of the robot are zero, the high correlation between the two
signals demonstrates the effectiveness of leg odometry drift rejection.

\subsection{Terrain Reconstruction Assessment}
\begin{figure}
\centering
\includegraphics[height=3.5cm]{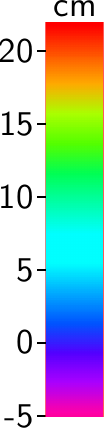}
\includegraphics[height=3.5cm]{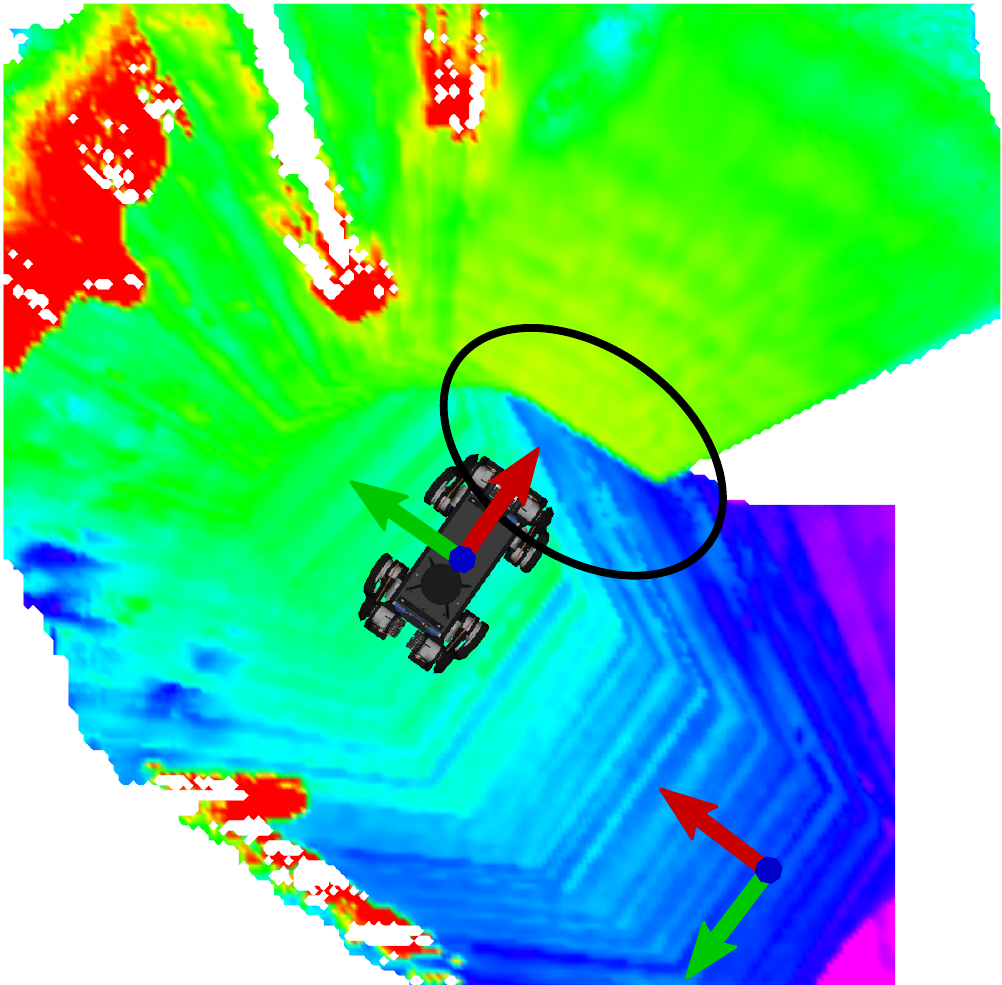}%
\hspace{0.01\columnwidth}%
\includegraphics[height=3.5cm]{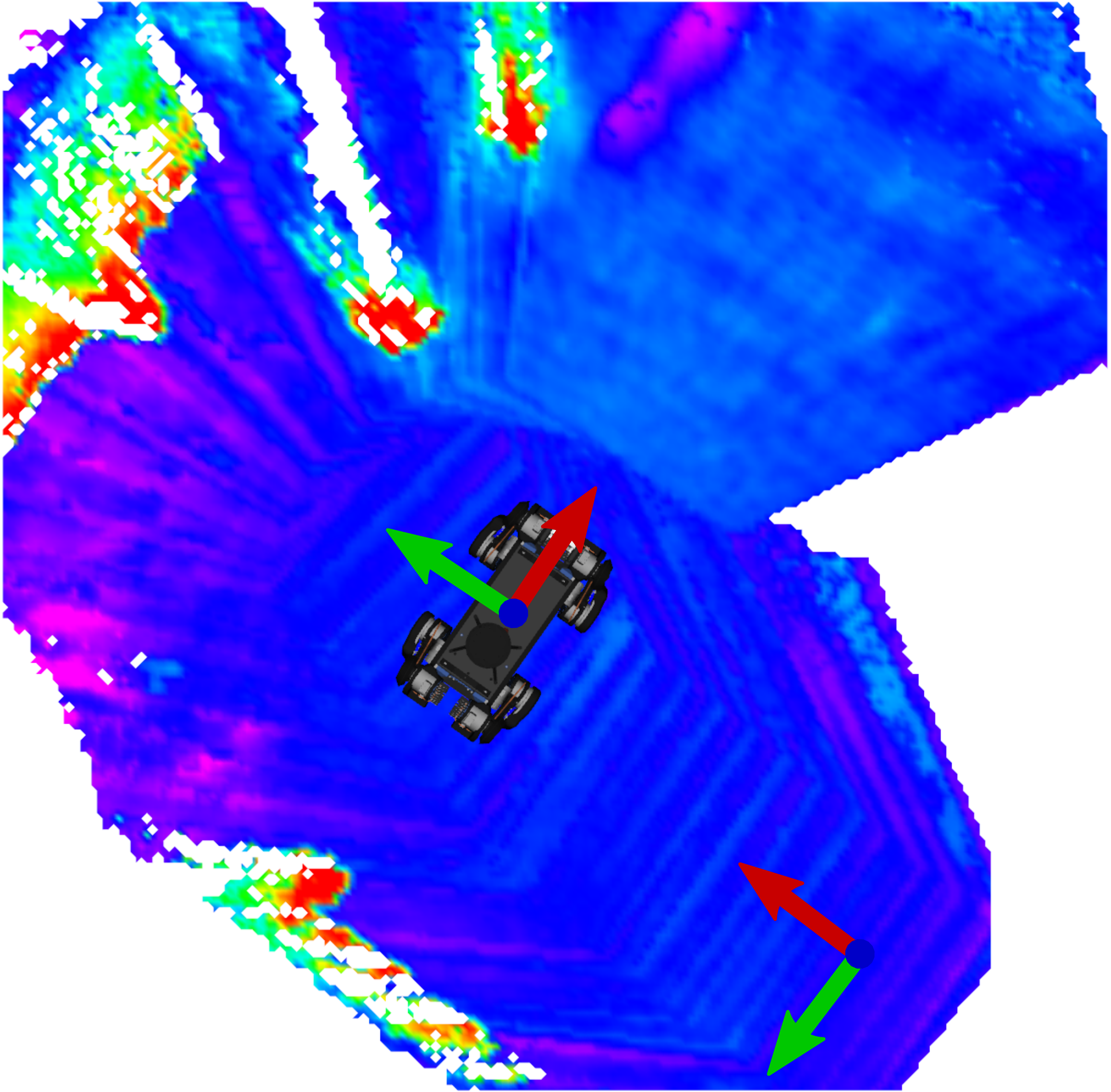}
\caption{Terrain reconstruction comparison between TSIF \cite{Bloesch2017}
(left) and VILENS (right). The robot walks from the bottom right corner to the
center of the image and turns \SI{90}{\degree} right. VILENS eliminates the
\SI{15}{\centi\meter} drift and drastically reduces the number of artifacts on
the elevation map.}
\label{fig:elevation-mappings}
\vspace{-5mm}
\end{figure}

We have evaluated the quality of local terrain mapping during a sequence of
walking and turning on flat ground from the FSC dataset (Fig.
\ref{fig:elevation-mappings}). Due to drift in the ANYmal's internal filter
\cite{Bloesch2017}, the elevation map contains a phantom discontinuity in front
of the robot (encircled in black). With VILENS, the drift is reduced for
effective footstep planning.

\section{Conclusion}
\label{sec:conclusions}
We have presented a novel factor graph formulation for state estimation that
estimates preintegrated velocity factors for leg odometry and velocity bias
estimation to accommodate for leg odometry drift. These bias effects are
difficult to directly model, we instead infer them from vision. The redundancy
of our approach is also demonstrated in visual impoverished situations which
vision along would struggle. In these situations, our system gracefully relies
on leg odometry and the velocity bias estimation compensates for its drift. We
have demonstrated the robustness of our method with outdoor experiments which
include conditions such as slippery and deformable terrain, reflections, and
external disturbances applied to the robot.

\section{Acknowledgements}
This research has been conducted as part of the ANYmal research community. It
was part funded by the EU H2020 Projects THING and MEMMO, a Royal Society
University Research Fellowship (Fallon) and a Google DeepMind studentship
(Wisth). Special thanks to Ruben Grandia and Matthew Jose Pollayil for the
support during experiments and the RSL group (ETH) for general support.

\bibliographystyle{./IEEEtran}
\bibliography{./IEEEabrv,./library}
\end{document}